\documentclass[10pt,twocolumn,letterpaper]{article}
\pdfoutput=1
\usepackage[final]{cvpr}      
\usepackage{times}
\usepackage{epsfig}
\usepackage{graphicx}
\usepackage{adjustbox}
\usepackage[ruled,linesnumbered]{algorithm2e}
\usepackage{amsmath}
\usepackage{amssymb}
\usepackage{booktabs} 
\usepackage{multirow} 
\usepackage{siunitx} 
\usepackage[accsupp]{axessibility} 
\usepackage{diagbox}
\usepackage{colortbl}
\usepackage{overpic}
\usepackage{bbm}
\usepackage{dsfont}
\definecolor{mygray}{gray}{.9}
\definecolor{white}{rgb}{1, 1, 1}

\newcolumntype{x}[1]{>{\centering\arraybackslash}p{#1pt}}

\usepackage[pagebackref,breaklinks,colorlinks]{hyperref}

\usepackage[capitalize]{cleveref}
\crefname{section}{Sec.}{Secs.}
\Crefname{section}{Section}{Sections}
\Crefname{table}{Table}{Tables}
\crefname{table}{Tab.}{Tabs.}

\def\cvprPaperID{3912} 
\def\confName{CVPR}
\def\confYear{2022}

\begin{document}

\title{Back to Reality: Weakly-supervised 3D Object Detection with Shape-guided Label Enhancement}

\author{Xiuwei Xu\textsuperscript{1,2}, ~Yifan Wang\textsuperscript{1}, ~Yu Zheng\textsuperscript{1,2}, ~Yongming Rao\textsuperscript{1,2}, ~Jie Zhou\textsuperscript{1,2}, ~Jiwen Lu\textsuperscript{1,2}\thanks{Corresponding author.}\\
\textsuperscript{1}Department of Automation, Tsinghua University, China \\
~\textsuperscript{2}Beijing National Research Center for Information Science and Technology, China \\
{\tt\small \{xxw21, yifan-wa21, zhengyu19\}@mails.tsinghua.edu.cn;} \tt\small raoyongming95@gmail.com; \\
 {\tt\small \{jzhou, lujiwen\}@tsinghua.edu.cn} \\
}

\maketitle

\begin{abstract}
In this paper, we propose a weakly-supervised approach for 3D object detection, which makes it possible to train a strong 3D detector with position-level annotations (i.e.\ annotations of object centers).
In order to remedy the information loss from box annotations to centers, our method, namely Back to Reality (BR), makes use of synthetic 3D shapes to convert the weak labels into fully-annotated virtual scenes as stronger supervision, and in turn utilizes the perfect virtual labels to complement and refine the real labels.
Specifically, we first assemble 3D shapes into physically reasonable virtual scenes according to the coarse scene layout extracted from position-level annotations. Then we go back to reality by applying a virtual-to-real domain adaptation method, which refine the weak labels and additionally supervise the training of detector with the virtual scenes.
Furthermore, we propose a more challenging benckmark for indoor 3D object detection with more diversity in object sizes for better evaluation.
With less than 5\% of the labeling labor, we achieve comparable detection performance with some popular fully-supervised approaches on the widely used ScanNet dataset.
Code is available at: \href{https://github.com/wyf-ACCEPT/BackToReality}{\emph{https://github.com/wyf-ACCEPT/BackToReality}}.

\end{abstract}

\section{Introduction}
3D object detection is a fundamental scene understanding problem, which aims to detect 3D bounding boxes and semantic labels from a point cloud of 3D scene. Due to the irregular form of point clouds and complex contexts in 3D scenes, most existing 2D methods~\cite{ren2016faster,redmon2016you,zhou2019objects} cannot be directly applied to 3D object detection. Fortunately, with the development of deep learning techniques on point cloud understanding~\cite{qi2017pointnet,qi2017pointnet++}, recent works~\cite{shi2018pointrcnn,zhou2018voxelnet,hou20193d,Qi_2019_ICCV,liu2021group} have employed deep neural networks to directly detect objects from point clouds and achieved favorable performance.

\begin{figure}
\centering
\vspace{-2pt}
\includegraphics[width=1.0\linewidth]{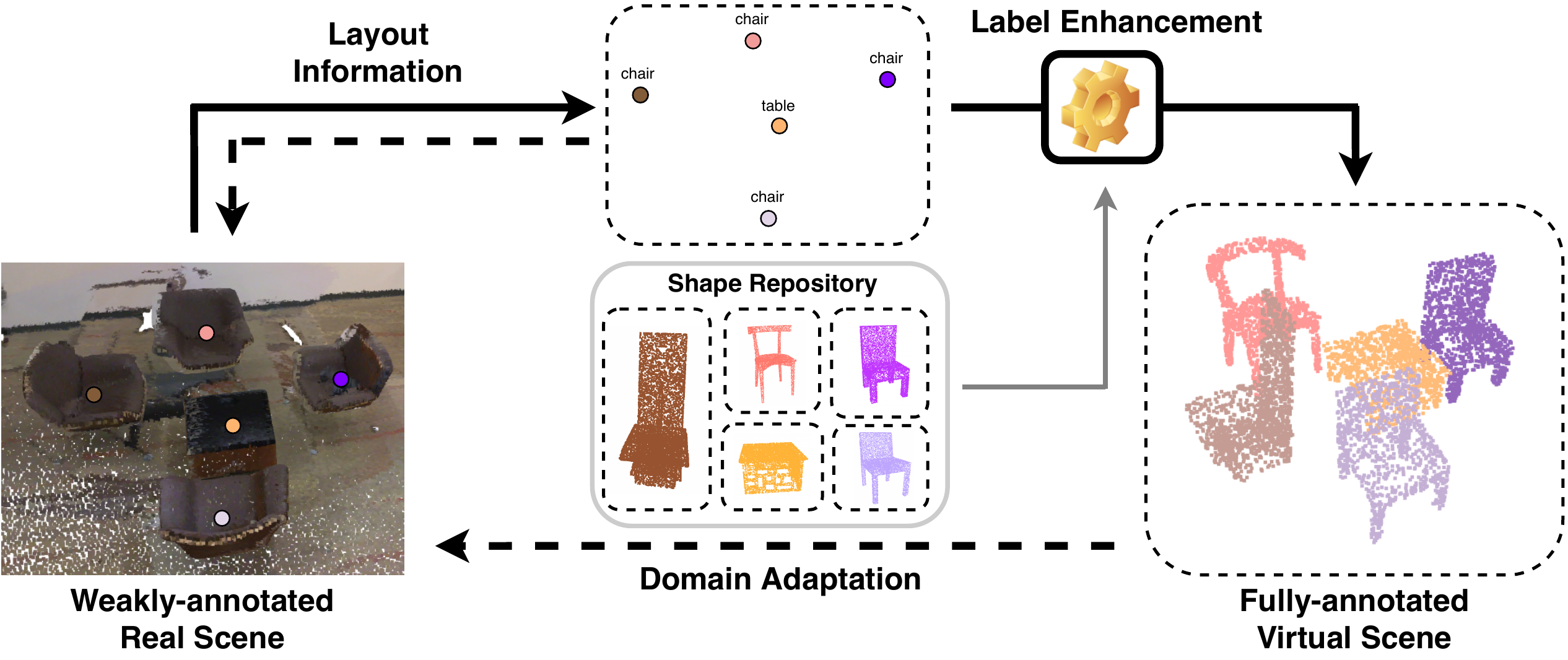}
\vspace{-18pt}
\caption{Demonstration of \textit{BR}. We consider position-level annotations as the coarse layout of the scenes, which is utilized to generate virtual scenes from a 3D shape repository. Physical constraints are applied on the virtual scenes to remedy the information loss from box annotations  to centers. Then a virtual-to-real domain adaptation method is presented to additionally supervise the real-scene 3D object detection with the virtual scenes. Dashed arrows indicate supervision for training.}
\vspace{-1mm}
\label{fig:toy_example}
\end{figure}

Despite the successes in deep learning based object detection on point clouds, massive amounts of labeled bounding boxes are required for training the detector. This issue significantly limits the applications of these methods, as labeling a precise 3D box takes more than 100s even by an experienced annotator~\cite{song2015sun}. Therefore, 3D object detection methods using cheap labels are desirable for practical applications. 
Motivated by this, increasing attention has been paid to weakly-supervised 3D object detection methods, which can be divided into two categories according to the form of annotation: scene-level~\cite{ren20213d} and position-level~\cite{meng2020weakly,meng2021towards} where only the class tag and both object center and class are annotated for each object respectively. The two types of annotation only require less than 1\% and 5\% time for one instance compared to labeling a bounding box, as shown in Table \ref{annotate}. While scene-level annotation is more time-saving, it is hard for the detector to learn how to precisely locate each object in a scene due to the lack of position information, and thus the performance is far from satisfactory~\cite{ren20213d}. Considering the time-accuracy tradeoff, position-level annotation is a more practical solution. However, previous position-level weakly-supervised 3D detection methods still require a number of precisely labeled boxes and can only cope with sparse outdoor scenes~\cite{meng2020weakly,meng2021towards}. Purely position-level weakly-supervised method for the complicated indoor detection task is still under exploration.

In this paper, we propose a shape-guided label enhancement approach called \textit{Back to Reality} (BR) for weakly-supervised 3D object detection\footnote{Label enhancement (LE) is a technique to recover label distributions from logical labels, as defined in \cite{xu2019label}. Here we extend the concept of LE to denote the process of recovering the lost information for weak labels.}. To reduce the labor cost, we only label the center of each object in the 3D space and the labeling error of centers is allowed\footnote{We show the detailed labeling strategy in Section \ref{annotation}.}. 
While largely reducing the workload of labeling, the information loss is non-negligible from box annotations to centers. To address these, BR converts the weak labels into virtual scenes which contain much of the lost information, and in turn utilizes them to additionally supervise real-scene training, as shown in Figure \ref{fig:toy_example}. Our approach is based on two motivations: 1) in 3D vision, large-scale datasets of synthetic shapes are available. They contain rich geometry information, which can serve as strong prior to assist 3D object detection; 2) the position-level annotations are not only supervision for training, but they also provide coarse layout of the scene. Therefore, we assemble the 3D shapes into fully-annotated virtual scenes according to the coarse layout and apply physical constraints on them to remedy the information loss. 
Then a virtual-to-real domain adaptation method is presented to align the global features and object proposal features extracted by the detector between the real and virtual scenes. 
Moreover, our method can take advantage of the precise center labels in virtual scenes to correct the center error of position-level annotations. 
In this way the useful knowledge contained in virtual scenes is transferred back to reality. Experimental results on ScanNet~\cite{dai2017scannet} show the effectiveness of the proposed BR method.

\begin{table}[]
    \centering
    \setlength\tabcolsep{5pt}
    \caption{Annotating time and detection results of different methods based on various types of annotation. The benchmark is detailed in Section \ref{exp}. (BBox refers to box annotation. S-L and P-L mean scene-level and position-level annotations respectively.)}
    \vspace{-1mm}
    \footnotesize
    \begin{tabular}{|c|p{1.1cm}<{\centering}|p{1.1cm}<{\centering}|p{1.1cm}<{\centering}|p{1.1cm}<{\centering}|}
        \hline
        Annotation & BBox~\cite{liu2021group} & S-L~\cite{ren20213d} & P-L~\cite{meng2020weakly} & P-L(BR) \\
        \hline
        Time(s per object) & 110 & 1 & 5 & 5 \\
        mAP@0.25(\%) & 54.2 & \textless 20 & 32.4 & 47.0 \\
        \hline
    \end{tabular}
    \label{annotate}
    \vspace{-1mm}
\end{table}

\section{Related Work}

\textbf{3D Shape to Scene:}
Since it is much easier to obtain a large scale synthetic 3D shape dataset than a real scene dataset, utilizing the shapes to assist scene understanding is a promising idea. Existing approaches can be divided into two categories: supervised~\cite{avetisyan2019scan2cad,avetisyan2019end,dahnert2019joint,uy2020deformation} and unsupervised~\cite{dosovitskiy2015flownet,wang2019normalized,Liu_2019_CVPR,peng2020convolutional,rao2021randomrooms}. For supervised methods, the synthetic shapes are usually used to complete the imperfect real scene scans. 
Given a set of CAD models and a real scan, a network is trained to predict how to place the CAD models in the scene and replace the partial and noisy real objects~\cite{avetisyan2019scan2cad,avetisyan2019end,dahnert2019joint,uy2020deformation}. Human-annotated pairs of raw scans and object-aligned scans are used in the training process. 
As supervised methods need extra human labor, that may limit the full utilization of 3D shape datasets.    
Unsupervised methods are usually used for data augmentation or dataset expansion. 3D CAD models are placed in a random manner following the basic physical constraints, in order to generate mixed reality scenes~\cite{dosovitskiy2015flownet,wang2019normalized} or virtual scenes~\cite{Liu_2019_CVPR,peng2020convolutional}.
Recently, RandomRooms~\cite{rao2021randomrooms} proposes to use ShapeNet dataset for unsupervised pretraining of 3D detector. Our approach also utilizes 3D shapes to assist object detection in an unsupervised manner. Differently, we aim to make use of synthetic shapes to enhance the weak label and gain stronger supervision in position-level weakly-supervised detection task.


\textbf{3D Object Detection:}
Early 3D object detection methods mainly include template-based methods~\cite{li2015database,litany2017asist,nan2012search} and sliding-window methods~\cite{song2014sliding,song2016deep}. Deep learning-based 3D detection methods for point clouds began to emerge thanks to PointNet/PointNet++~\cite{qi2017pointnet,qi2017pointnet++}. 
However, methods in \cite{qi2018frustum,chen2017multi,chen2016monocular,lahoud20172d} rely on generating 2D proposals and then project them into the 3D space, which is hard to handle scenes with heavy occlusion. 
More recently, networks that directly consume point clouds have been proposed~\cite{shi2018pointrcnn,zhou2018voxelnet,hou20193d,Qi_2019_ICCV,liu2021group}. 
While the development of 3D object detection methods is rapid, the application is still restricted partially due to the limited labeled data. To reduce the labor of human annotation, weakly-supervised methods~\cite{qin2020weakly,meng2021towards,meng2020weakly,ren20213d}, semi-supervised methods~\cite{wang20213dioumatch,zhao2020sess} and unsupervised pretraining methods~\cite{xie2020pointcontrast,hou2020exploring,zhang2021self,rao2021randomrooms} have been proposed recently. 
However, pretraining methods rely on huge computing resources for training the networks in a contrastive learning manner. Semi-supervised methods follow the similar procedure as their 2D counterparts~\cite{tarvainen2017mean} and do not fully exploring the characteristics of 3d data. Therefore, we investigate weakly-supervised approach tailored for 3D object detection task.

\section{Approach}
\begin{figure*}
\centering
\vspace{-2mm}
\includegraphics[width=1.0\linewidth]{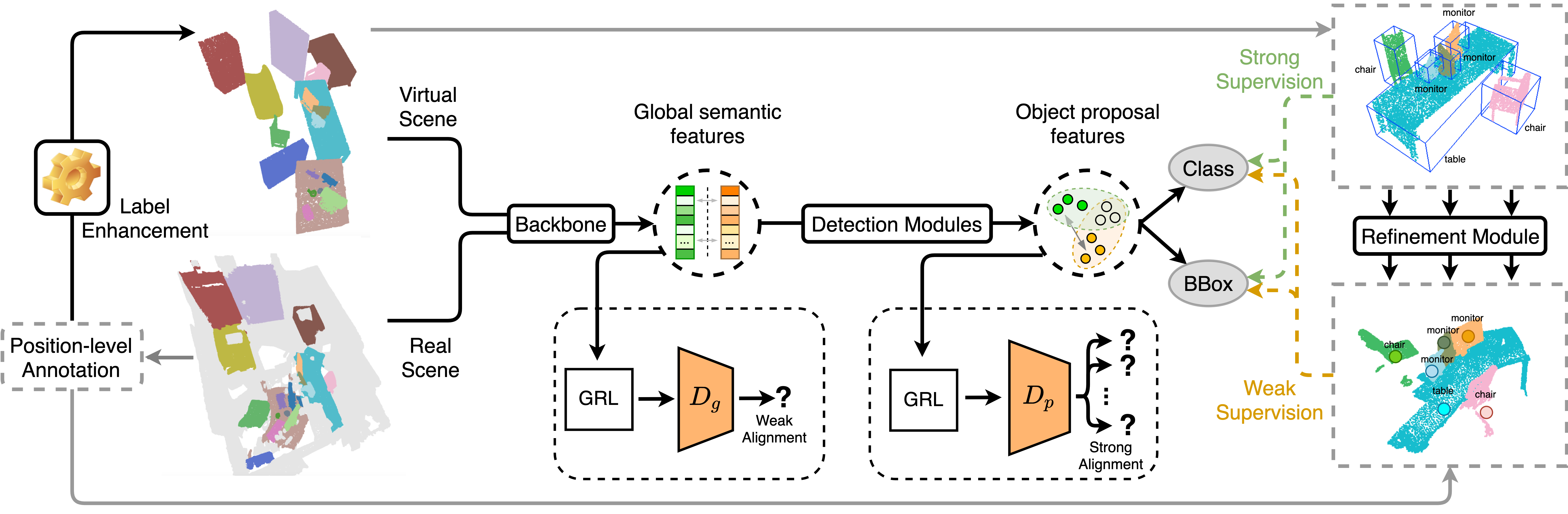}
\vspace{-7mm}
\caption{	
The framework of our \textit{BR} approach. Given real scenes with position-level annotations, we first enhance the weak labels to get fully-annotated virtual scenes. Then the real scenes and virtual scenes are fed into the detector, trained with weakly-supervised and fully-supervised detection loss respectively. During training we use the precise object centers in virtual scenes to refine the imprecise centers in real scenes. Strong-weak adversarial domain adaptation method is utilized to align the distributions of features from both domains. The global discriminator outputs judgments for each scene, and the proposal discriminator outputs judgments for each object proposal. (Here GRL refers to gradient reversal layer; $D_g$ and $D_p$ stand for the global and proposal discriminators respectively.)
}
\vspace{-2mm}
\label{fig:method_overview}
\end{figure*}

Figure \ref{fig:method_overview} illustrates the framework of our approach. Given real scenes with position-level annotations, we utilize 3D shapes to convert the weak labels into virtual scenes, which are utilized to provide additional supervision for the training of the detector. In this section, we first discuss our weakly-supervised setting and then demonstrate the key steps of BR.


\subsection{Position-level Annotation}\label{annotation}
As choosing a point in the 3D space is hard, we divide the labeling process into two steps: firstly we label the center of an object in a proper 2D view of the scene, and compute the line that goes through this center and the focus point of the camera according to the camera parameters of the 2D view. Secondly we choose a point on the line to determine the object's center in the 3D space. This strategy requires less than 5s to label an instance, and the labeling error can be controlled within 10\% of the instance size.

When the 3D scene is scanned, in many cases we can acquire \textbf{mesh} data. We assume the meshes are available in our input. Nevertheless, case where we only have point cloud data is also considered in our approach and experiments.

\subsection{Shape-guided Label Enhancement}\label{subsec3}


While position-level annotation requires far less labeling time, its information loss is severe, which is manifested in two aspects: 1) the information of objects' sizes is lost; 2) the object centers are imprecise. In spite of this, position-level annotations can provide a coarse layout of the scenes. By assembling synthetic 3D shapes according to the layout, we are able to enhance the weak labels and generate accurately-annotated virtual scenes where sizes are available and centers are precise.
Our label enhancement method is two-step: 1) first we calculate some basic properties of 3D shapes; 2) then we place these shapes to generate physically reasonable virtual scenes from the labels. We provide some implementation details in the supplementary\footnote{we use $^*$ to indicate that the exact definition is in supplementary.}.

\textbf{Definition of Shape Properties: } Given a synthetic 3D shape, which is represented as $O\in R^{N\times 3}$, we assume it is axis-aligned and normalized into a unit sphere. The length, width and height of $O$ is defined as $l$, $w$ and $h$. Then we divide the categories of shapes into three classes: supporter, stander and supportee. Supporters and standers are objects that can only be supported by ground, with the difference that standers are not likely to support other things. Other categories are supportees.

Then if a shape belongs to supporter, three properties are calculated: minimum-area enclosing rectangle ($MER^*$), supporting surface height ($SSH^*$) and compactness of the supporter surface ($CSS^*$). The $MER$ is computed in XY plane, which is the minimum rectangle enclosing all the points of the shape. The $SSH$ is the height of the highest surface on which other objects can stand. The $CSS$ is a boolean value, indicating whether the supporting surface can be approximated by the $MER$.

\begin{figure*}
\centering
\vspace{-2mm}
\includegraphics[width=1.0\linewidth]{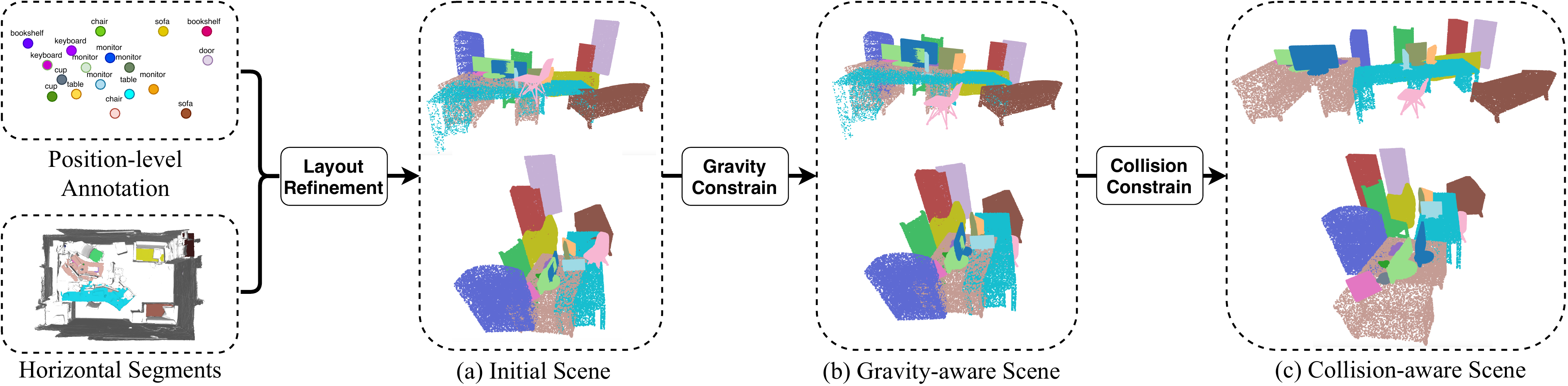}
\vspace{-6mm}
\caption{The pipeline of our three-stage virtual scene generation method. We first extract horizontal segments from the mesh data and use them to refine the coarse layout provided by position-level annotations. Then synthetic 3D shapes are placed in virtual scenes according to the new layout to construct initial virtual scenes. After that we apply gravity and collision constraints on the virtual scenes to restore the lost physical relationships between objects and make the scenes more realistic.}
\vspace{-2mm}
\label{fig:generation}
\end{figure*}

\textbf{Virtual Scene Generation: } We utilize a three-stage approach to construct the virtual scenes, which is equivalent to generate the position of each shape stage by stage: 1) we first refine the coarse layout provided by position-level annotations and generate the initial positions; 2) then we generate gravity-aware positions by restoring the supporting relationships between objects; 3) lastly we generate collision-aware positions to make the virtual scenes physically reasonable. The pipeline is shown in Figure \ref{fig:generation}.

To generate \textit{initial positions}, we need to recover a more precise layout from the geometric information of the scenes.
Given a scene in mesh format, we first oversegment the meshes using a normal-based graph cut method~\cite{felzenszwalb2004efficient,karpathy2013object}. The result is a segment graph, where the nodes indicating segments and the edges denoting adjacency relations. Then for horizontal$^*$ segments whose area$^*$ is larger than $A_{min}$ and height$^*$ is larger than $H_{min}$, we iteratively merge their neighbors into them if the height difference between the horizontal segment and the neighbor segment is smaller than $\Delta_h$. Once merged, the segments are considered as a whole and the height of the new merged segment is set to be same as the original horizontal segments. After merging, each horizontal segment is represented by its $MER$. If only one supporter's center falls in a $MER$, we assign this $MER$ to the supporter. When the centers of multiple supporters fall in the same $MER$, we perform K-means clustering of the horizontal segment according to these centers and calculate $MER$ for each supporter respectively.

Then we place the 3D shapes of corresponding categories on the centers given by position-level annotations and utilize the horizontal segments to refine the layout.
The initial positions of the shapes are represented by a dictionary, whose key is the instance index and value is a list:
\begin{equation}\label{value}
    [(x, y, z), (s_x, s_y, s_z), O, \theta, S, M, H]
\end{equation}
where the instance index is a integer ranging from 1 to the number of objects in the scene. $(x, y, z)$ denotes the center coordinates. $(s_x, s_y, s_z)$ indicates the scales in three dimensions. $\theta$ is the rotation angle of the shape. $S$ tells whether the shape is a supporter. $M$ and $H$ indicate the $MER$ and $SSH$ of supporter. They are set to None when $S$ is false. If the shape has been assigned a horizontal segment, we use the $MER$ of that segment to initialize the above parameters. That is, we choose a supporter whose $CSS$ is True and make the $MER$ of this supporter overlap with the horizontal segment.
Otherwise we conduct random initialization. If only point cloud data is available, we simply perform random initialization and the following stages are the same.

Next we traverse the initial positions to generate \textit{gravity-aware positions}. In this process we only need to change $z$ and $SSH$ in the position dictionary. For supporters and standers, we directly align their bottoms with the ground (i.e.\ the XY plane). For a supportee, if its $(x, y)$ fall in any supporter's $MER$, we assign it to the nearest supporter and align its bottoms with the supporting surface. Otherwise, it is aligned to the ground. 

After that we move the shapes to acquire \textit{collision-aware positions}. This stage only $x$ and $y$ in the position dictionary will be changed. First we move the objects on the ground, the supported ones on which will move together if there are. Then for each supporter, we move its supportees until there is no overlap. Note that the three generation stages can not only make the virtual scenes more realistic, but also weaken the impact of imprecise center labels. Thus the virtual scene generation method is robust to labeling errors.

Finally, we convert the collision-aware positions to point clouds with proper density. As larger surfaces are more likely to be captured by the sensor, we use the maximum of $(ls_x)(ws_y)$, $(ws_y)(hs_z)$ and $(ls_x)(hs_z)$ to approximate the surface area of shapes. Then the number of points for each object is set proportional to their surface areas using uniform sampling, the largest one remaining $N$ points.

\subsection{Virtual2Real Domain Adaptation}\label{subsec4}
While the label enhancement approach is able to generate physically reasonable fully-annotated virtual scenes, there is still a huge domain gap between them and the real scenes (e.g.\ backgrounds like walls are missed in the virtual scenes). Therefore, we need to mining useful knowledge in the perfect virtual labels to make up for the information loss of position-level annotations, rather than just relying on the virtual scenes.

We refer to the virtual scenes and real scenes as source domain and target domain respectively. A virtual-to-real adversarial domain adaptation method is utilized to solve the above problem, whose overall objective is:
\begin{equation}
    \begin{split}
        \max\limits_D\min\limits_O J&=L_{sup}(O)-L_{adv}(O,D)\\
        &=(L_{1}+L_{2}+L_{3})-(L_{4}+L_{5})
    \end{split}
\end{equation}
where $O$ refers to the object detection network (detector) and $D$ indicates the discriminators used for adversarial feature alignment. $L_{sup}$ aims to minimize the differences between the predicted bounding boxes and the annotations, which can be further divided into the loss for center refinement module ($L_1$), fully-supervised detection loss on source domain ($L_2$) and weakly-supervised detection loss on target domain ($L_3$). The objective of $L_{adv}$ is to align the features from source domain and target domain, which aims to utilize the knowledge learned from source domain to assist object detection in target domain. $L_{adv}$ can be divided into global feature alignment loss ($L_4$) and proposal feature alignment loss ($L_5$). Below we will explain these loss functions and our network in detail.

Firstly we elaborate on $L_{sup}(O)$. 
As shown in Figure \ref{fig:method_overview}, we divide the detector into three blocks: a backbone which extracts global semantic features from the scene, a detection module which generates object proposals from the semantic features, and a prediction head which predicts the semantic label and bounding box from each object proposal feature. 

\begin{figure}
	\centering
	\includegraphics[width=1.0\linewidth]{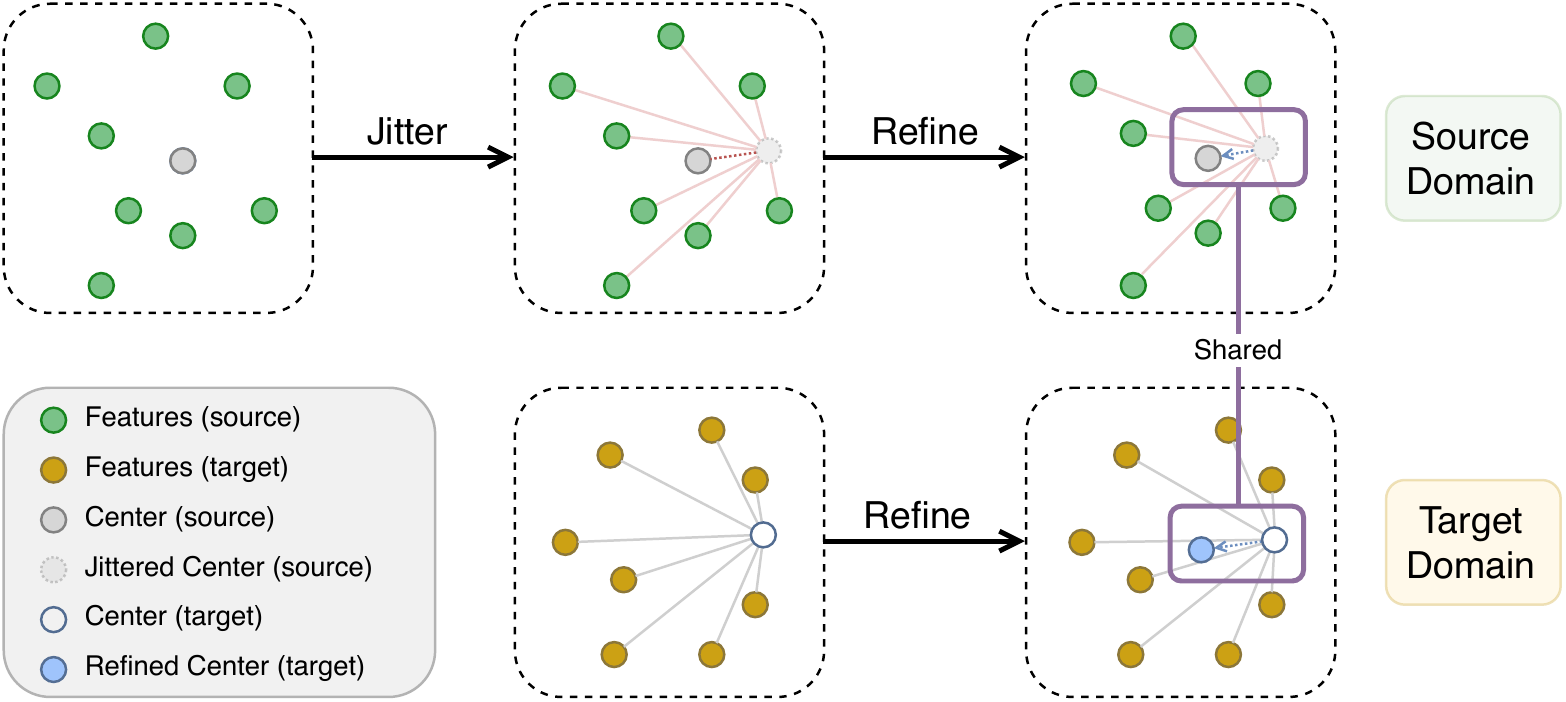}
	\vspace{-18pt}
	\caption{Demonstration of our center refinement method. We first jitter the center labels in source domain, and utilize a PointNet-like module to predict the center offset from the local graph of the jittered centers. This module can be directly utilized to predict the center error in target domain as the global semantic features from the two domains have been aligned.}
	\vspace{-12pt}
	\label{fig:ctjt}
\end{figure}

During training, we jointly refine the imprecise center labels in target domain and supervise the predictions of the detector. As shown in Figure \ref{fig:ctjt}, we jitter the center labels in source domain by adding noise within 10\% of the objects' sizes to imitate the labeling error in target domain. Then for each jittered center, we query its $k$ nearest neighbors in 3D euclidean space from the global semantic features to construct a local graph, and predict the center offset through a PointNet-like module:
\begin{equation}\label{ctjt}
    p(c) = \text{MLP}_2 \left\{ \underset{i\in N(c)}{\mbox{max}}\left\{\text{MLP}_1 [f_i; c_i-c]\right\} \right\}
\end{equation}
where $p$ denotes the PointNet-like module, $c$ indicates the jittered center label, $N(c)$ is the index set of the $k$ nearest neighbors of $c$, $f_i$ is the global semantic feature, whose coordinate is $c_i$, and \textit{max} refers to the channel-wise max-pooling. 
We set $\boldsymbol{L_1}$ as the mean square error between the ground-truth center offset and $p(c)$. 
Then for fully-supervised training, the detection loss $\boldsymbol{L_2}$ is the same as the loss utilized in the original method. For weakly-supervised training, we utilize $p$ to predict the center error in target domain and acquire refined center labels. We set $\boldsymbol{L_3}$ as a simpler version of $L_2$ which ignores the supervision for box sizes. More details about $L_3$ can be found in supplementary.

Secondly we analyze $L_{adv}(O,D)$. We conduct feature alignment in an adversarial manner: the discriminator predicts which domain the features belong to, and the detector aims to generate features that are hard to discriminate. The sign of gradients is flipped by a gradient reversal layer~\cite{ganin2015unsupervised}.

As the virtual scenes and real scenes are processed by the same network, we hope $L_3$ helps the network learn how to locate each object in real scenes, and $L_2$ compensates for the information loss of centers and sizes. However, due to the domain gap, $L_2$ will introduce domain-specific knowledge of the virtual scenes, which impairs the influence of $L_3$. Besides, the center refinement module is trained only on source domain, which may not perform well on target domain. Therefore, we align the global semantic features and object proposal features with $L_4$ and $L_5$ respectively. 
Inspired by \cite{saito2019strong}, the features are aligned with different intensities at different stages. 
For global semantic features, we use a PointNet to predict the domain label. Focal loss~\cite{saito2019strong,lin2017focal} is utilized to apply weak alignment:
\begin{equation}\label{focal}
    \boldsymbol{L_4}=-\sum_{i=1}^B{(1-p_i)^\gamma log(p_i)},\  \gamma>1
\end{equation}
where $B$ is the batch size, and $p_i$ refers to the probability of the global discriminator's predictions on the corresponding domain. Features with high $p$ is easy to judge, which means they are domain-specific features and forcing invariance to them can hurt performance. So a small weight is used to reduce their impact on training. For object proposal features, they will be directly taken to predict the properties for bounding boxes. As the properties are domain-invariant and have real physical meaning, we strongly align this stage of features using an objectness weighted L2 loss:
\begin{equation}
    \boldsymbol{L_5}=\sum_{i=1}^B{\sum_{j=1}^N{s_{ij}(1-p_{ij})^2}}
\end{equation}
where $B$ is the batch size, $N$ is the number of proposals, $s_{ij}$ refers to the objectness label and $p_{ij}$ is the probability of the proposal discriminator's predictions on the corresponding domain. We detail the architectures of center refinement module and discriminators in supplementary.

\begin{table*}
	\caption{Number of objects in each category in the training set and validation set of ScanNet, and average number of points of objects in each category in the real scenes and the virtual scenes.}\label{tab0}
	\vspace{-1mm}
	\begin{adjustbox}{width=1\textwidth}
	{\LARGE\begin{tabular}{|l|c|c|c|c|c|c|c|c|c|c|c|c|c|c|c|c|c|c|c|c|c|c|c|}
	  \hline
	  & \multirow{2}{*}{Property} & Bath- & \multirow{2}{*}{Bed} & \multirow{2}{*}{Bench} & Book- & \multirow{2}{*}{Bottle} & \multirow{2}{*}{Chair} & \multirow{2}{*}{Cup} & Cur- & \multirow{2}{*}{Desk} & \multirow{2}{*}{Door} & \multirow{2}{*}{Dresser} & Key- & \multirow{2}{*}{Lamp} & \multirow{2}{*}{Laptop} & \multirow{2}{*}{Monitor} & Night- & \multirow{2}{*}{Plant} & \multirow{2}{*}{Sofa} & \multirow{2}{*}{Stool} & \multirow{2}{*}{Table} & \multirow{2}{*}{Toilet} & Ward- \\
	  &  & tub & & & shelf & & & &tain & & &  &board & & & &stand & & & & & &robe \\
	  \hline
	 \# train &\multirow{2}{*}{Object Number} &113 &308 &58 &786 &234 &4357 &132 &408 &551 &2028 &174 &193 &376 &86 &574 &190 &293 &406 &315 &1526 &201 &98 \\
	  \cline{1-1}\cline{3-24}
	 \# validate & &31 &81 &21 &234 &41 &1368 &34 &95 &127 &467 &43 &53 &83 &25 &191 &34 &50 &97 &51 &407 &58 &19 \\
	  \hline
	 \# real &\multirow{2}{*}{Point Number} &2941 &3905 &1015 &2679 &101 &726 &66 &2919 &1525 &1110 &1274 &74 &272 &173 &370 &700 &792 &2718 &525 &1282 &1445 &2762 \\
	  \cline{1-1}\cline{3-24}
	 \# virtual & &6891 &8683 &4097 &6258 &162 &2135 &91 &5495 &5004 &6048 &2703 &480 &609 &343 &939 &1088 &1249 &7250 &1391 &5421 &3716 &6105 \\
	  \hline
	\end{tabular}}
	\end{adjustbox}
	\vspace{-1mm}
\end{table*}

\begin{table*}[]
	\centering
	\setlength{\abovedisplayskip}{0pt}
	\setlength{\belowdisplayskip}{0pt}
	\footnotesize
	\setlength\tabcolsep{1.6pt}
	\caption{The class-specific detection results (mAP@0.25) of different weakly-supervised methods on ScanNet validation set. (FSB is the fully-supervised baseline. $^\dag$ indicates the method requires a small proportion of bounding boxes to refine the prediction. Other methods only use position-level annotations as supervision. We set best scores in bold, runner-ups underlined.)}
	\vspace{-1mm}
	\begin{tabular}{c|l|cccccccccccccccccccccc|c}
		\hline
		 & Setting &batht. &bed &bench & bsf. & bot. &chair & cup &curt. &desk &door & dres. & keyb. &lamp &lapt. & monit. & n.s. &plant & sofa & stool & table & toil. & ward. &mAP@0.25 \\
		\hline \hline
		\rowcolor{mygray}
		\cellcolor{white} \multirow{6}{*}{\rotatebox[origin=c]{90}{VoteNet}} & FSB~\cite{Qi_2019_ICCV} &66.8 &86.2 &\underline{24.4} &\textbf{55.6} &0.0 &\underline{88.3} &0.0 &\underline{48.5} &\underline{62.8} &\underline{45.8} &24.1 &0.1 &47.2 &5.2 &62.1 &73.2 &13.4 &\underline{88.7} &35.1 &\underline{62.6} &94.6 &7.8 &45.1 \\
		 & WSB &21.9 &46.9 &0.3 &2.3 &0.0 &53.7 &0.0 &0.9 &32.1 &1.0 &6.6 &0.1 &0.2 &0.1 &1.8 &53.6 &0.1 &57.0 &4.6 &6.4 &19.7 &0.0 &14.1 \\ 
		 & WS3D$^\dag$~\cite{meng2020weakly} &22.0 &58.5 &10.3 &5.8 &0.0 &60.4 &0.0 &4.1 &26.7 &3.2 &1.6 &0.0 &14.0 &0.6 &18.6 &46.3 &0.4 &32.7 &11.8 &23.5 &65.0 &0.0 &18.4 \\
		 & WSBP$_P$ &43.2 &58.0 &2.4 &16.1 &0.0 &75.1 &0.7 &7.9 &54.2 &6.4 &7.1 &2.3 &35.2 &18.4 &12.8 &64.0 &4.4 &68.5 &20.2 &22.0 &71.6 &5.2 &27.1 \\ %
		 & WSBP$_M$ &45.0 &49.6 &5.5 &18.5 &0.0 &62.7 &2.9 &11.4 &49.6 &6.9 &2.5 &1.0 &30.0 &7.6 &21.4 &64.8 &7.3 &79.6 &23.1 &35.2 &80.9 &2.2 &27.6 \\
		 & BR$_P$(Ours) &51.2 &73.0 &16.4 &27.1 &0.1 &70.3 &0.0 &8.3 &44.5 &7.3 &16.0 &1.5 &40.2 &7.7 &42.1 &50.8 &7.4 &67.1 &10.7 &39.0 &88.4 &\underline{18.1} &31.2 \\
		 & BR$_M$(Ours) &57.1 &80.4 &14.3 &31.7 &0.0 &77.4 &0.0 &13.2 &49.7 &11.3 &14.8 &1.0 &43.5 &6.0 &56.5 &65.0 &10.6 &80.2 &26.9 &44.2 &91.4 &6.5 &35.5 \\
		\hline
		\rowcolor{mygray}
		\cellcolor{white} \multirow{6}{*}{\rotatebox[origin=c]{90}{GroupFree3D}} & FSB~\cite{liu2021group} &\textbf{86.2} &\underline{87.5} &16.3 &\underline{49.6} &0.6 &\textbf{92.5} &0.0 &\textbf{70.9} &\textbf{78.5} &\textbf{53.5} &\textbf{56.0} &6.4 &\textbf{68.2} &11.5 &\textbf{81.5} &\textbf{88.5} &15.2 &88.2 &\textbf{45.6} &\textbf{65.0} &\textbf{99.7} &\textbf{31.2} &\textbf{54.2} \\
		 & WSB &75.0 &75.7 &4.3 &17.2 &0.0 &81.4 &0.0 &3.5 &34.0 &4.7 &3.2 &2.1 &46.6 &3.3 &45.8 &52.8 &8.3 &71.0 &15.7 &18.1 &90.8 &0.7 &29.7 \\ 
		 & WS3D$^\dag$~\cite{meng2020weakly} &71.9 &78.3 &0.9 &20.2 &0.8 &79.2 &1.0 &2.9 &47.6 &7.7 &10.6 &\underline{19.2} &41.6 &13.5 &65.6 &41.2 &0.8 &74.6 &17.7 &26.3 &88.9 &1.7 &32.4 \\
		 & WSBP$_P$ &71.9 &77.1 &7.7 &25.2 &\textbf{3.0} &80.6 &0.4 &3.2 &50.1 &10.5 &36.3 &17.0 &52.9 &\textbf{30.3} &59.9 &63.8 &9.6 &78.2 &28.4 &25.3 &93.3 &14.4 &38.2 \\ %
		 & WSBP$_M$ &81.8 &82.6 &0.0 &35.0 &0.0 &77.5 &0.4 &27.1 &38.4 &7.6 &22.3 &9.7 &44.3 &24.4 &65.4 &76.5 &5.5 &62.4 &34.7 &28.7 &\textbf{99.7} &5.4 &37.7 \\
		 & BR$_P$(Ours) &72.3 &73.5 &\textbf{45.8} &27.7 &0.0 &77.2 &\textbf{8.2} &30.8 &35.0 &17.8 &\underline{51.7} &0.3 &64.2 &25.0 &63.5 &66.6 &\underline{23.8} &86.7 &33.9 &37.6 &98.3 &5.2 &43.0 \\
		 & BR$_M$(Ours) &\underline{85.3} &\textbf{90.9} &8.8 &34.3 &\underline{1.9} &80.0 &\underline{7.7} &24.7 &58.0 &20.8 &45.4 &\textbf{31.3} &\underline{64.4} &\underline{25.8} &\underline{67.5} &\underline{76.7} &\textbf{27.3} &\textbf{91.4} &\underline{43.3} &46.7 &94.8 &8.3 &\underline{47.1} \\
		\hline
	   \end{tabular}
	\label{tab1}
	\vspace{-2mm}
\end{table*}


\section{Experiment}\label{exp}
In this section, we conduct experiments to show the effectiveness of our BR approach. We first describe the datasets and experimental settings. Then we evaluate the generated virtual scenes and present the detection results of our method. We also design experiments to show the robustness of our virtual scene generation method and demonstrate the practicality of our approach. Finally we design several ablation studies to verify our scene generation and domain adaptation method.

\subsection{Experiments Setup}

\textbf{Datasets:}
We choose ModelNet40~\cite{wu20153d} as the dataset of synthetic 3D shapes. ModelNet40 contains 12,311 synthetic CAD models from 40 categories, split into 9,843 for training and 2,468 for testing. 
We perform experiments on the ScanNet~\cite{dai2017scannet} dataset. ScanNet is a richly annotated dataset of indoor scenes with 1201 training scenes and 312 validation scenes. 
For each object appeared in the scenes, ScanNet officially provides its corresponding class in ModelNet40.
Therefore we choose 22 categories of ModelNet40 which have more than 50 objects in ScanNet training set and 20 in the validation set, and report detection performance on them. Since ScanNet does not provide human-labeled bounding boxes, we predict axis-aligned bounding boxes and evaluate the prediction on validation set as in \cite{Qi_2019_ICCV,xie2020mlcvnet,zhang2020h3dnet,liu2021group}. We name this benchmark ScanNet-md40.

Compared to the 18-category setting in previous works~\cite{Qi_2019_ICCV,xie2020mlcvnet,liu2021group}, our ScanNet-md40 benchmark is more challenging. Apart from the categories of big objects (e.g.\ desk and bathtub), we also aim to detect relatively small objects, such as laptop, keyboard and monitor. Hence our benchmark can better evaluate the performance of both detectors and weakly-supervised learning methods.

\begin{figure*}
\centering
\vspace{-2mm}
\includegraphics[width=1.0\linewidth]{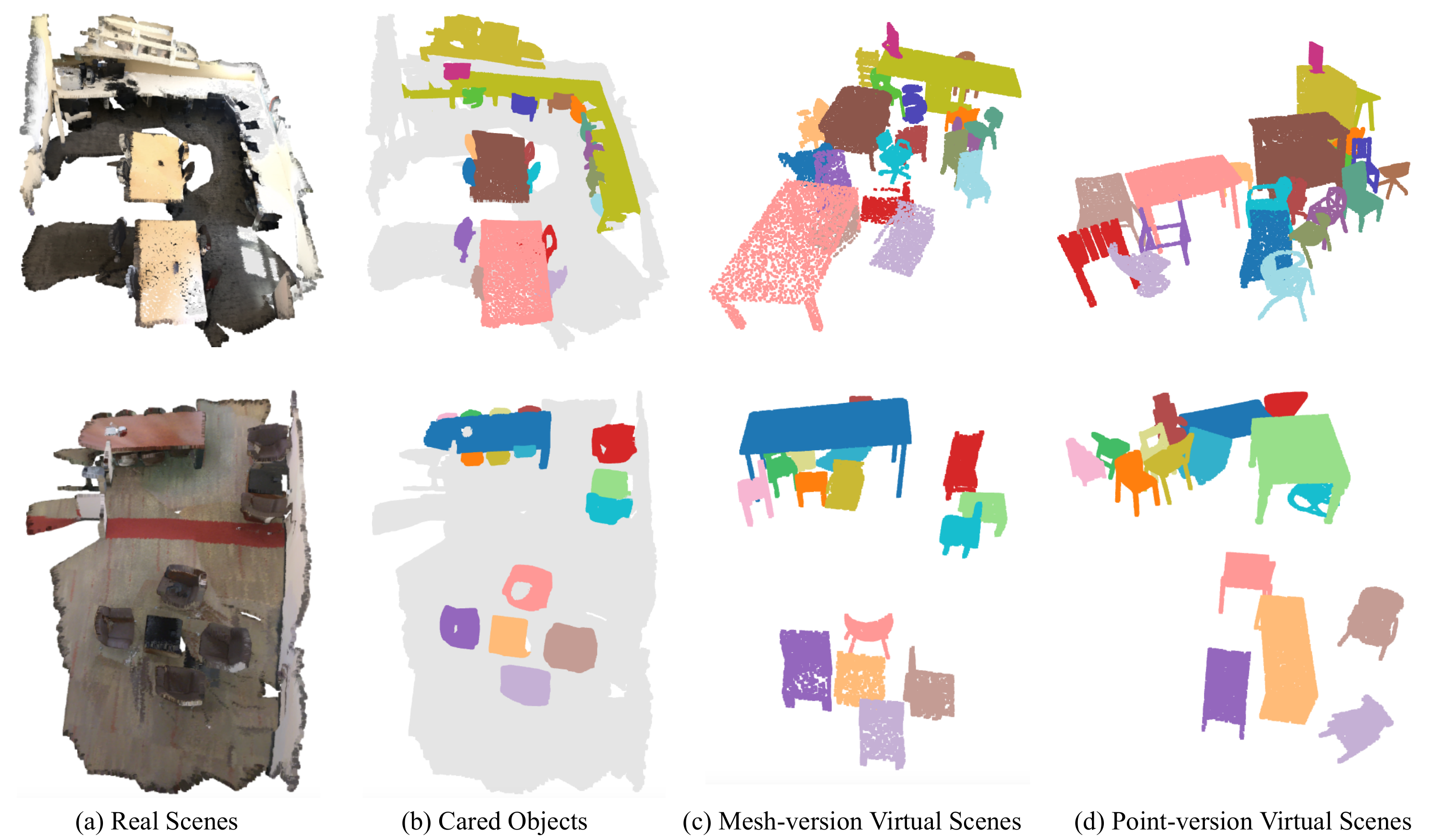}
\vspace{-18pt}
\caption{The qualitative visualization results of our virtual scene generation. In (b), (c) and (d), the same color indicates the same object. Gray points are floors, walls and objects that we do not care. It can be seen that the virtual scenes preserve the coarse scene context and the supporting relationships between objects.}
\vspace{-8pt}
\label{fig:quali}
\end{figure*}

\textbf{Compared Methods:}
To illustrate the effect of our BR approach, the popular VoteNet~\cite{Qi_2019_ICCV} and state-of-the-art GroupFree3D~\cite{liu2021group} are selected as our detectors. We compare BR with the following settings: 1) FSB: fully-supervised baseline, which serves as the upper bound of weakly-supervised methods; 2) WSB: weakly-supervised baseline, which trains the detector on real scenes by using $L_3$ only; 3) WS3D: another position-level weakly-supervised approach proposed in \cite{meng2020weakly}, which makes use of a number of precisely annotated bounding boxes; 4) WSBP: WSB pretrained on the virtual scenes. 
For settings which require the virtual scenes, we conduct experiments on two versions of virtual scenes (from points/meshes), which are distinguished by subscripts $M$ and $P$ respectively.

\textbf{Implementation Details:}
We set $N=10000$, $A_{min}=0.1 m^2$, $H_{min}=0.1m$, $\Delta_h=0.02m$, $k=16$ and $\gamma=3$. During training, as real scenes are more complicated, the converging of $L_3$ is much slower than $L_2$. Therefore we multiple $L_2$ by $0.1$ to slow down the training on virtual scenes and stabilize the process of feature alignment. To better train our center refinement module, the global semantic features should not change rapidly. Therefore we first train BR without $L_1$ until convergence, and then use the whole loss function to fine-tune the network. For GroupFree3D which has several decoders and each one outputs a stage of proposal features, we conduct feature alignment only for the last stage.

Different from previous works~\cite{Qi_2019_ICCV,liu2021group}, in our setting we need to detect small objects, such as bottle, cup and keyboard. As it is difficult for the network to extract high-quality features of these objects, we utilize an augmentation strategy to alleviate the problem, which is similar to \cite{kisantal2019augmentation}. 
Please refer to the supplementary for more details.


\subsection{Results and Analysis}
\textbf{Virtual Scene Evaluation:}
We first evaluate the statistics of the generated virtual scenes by computing the average number of points of objects in each category in real scenes and virtual scenes. As the input point clouds are downsampled to a given number before fed into the network, we only care about the ratio of average point numbers of objects in each category as the numbers can be controlled by the downsampling scale. We demonstrate the results in Table \ref{tab0}. It shows that the ratio in our virtual scenes is similar with that in the real scenes, which indicates the statistics of the virtual scenes are reasonable.

We also show qualitative visualizations to demonstrate our scene generation method in Figure \ref{fig:quali}. The virtual scenes generated with mesh information are named as mesh-version virtual scenes. Otherwise they are named as point-version virtual scenes. It is shown that the mesh-version virtual scenes can largely preserve the layout of the real scenes, and the point-version ones successfully combine the individual 3D shapes in a meaningful way.

\textbf{3D Object Detection Results:}\label{further}
As shown in Table \ref{tab1}, with position-level annotations only, WSB reduces the detection accuracy by a large margin in terms of mAP@0.25 compared to FSB. That's mainly because WSB fails to learn the ability of predicting precise centers and sizes of bounding boxes according to the scene context. 
WS3D makes use of some box annotations and achieve better performance. However, as it is specially designed for outdoor 3D object detection, WS3D is still far from satisfactory when coping with the complicated indoor scenes. 
With pretraining on the virtual scenes, WSBP has more than 8\% improvement over the WSB. That shows the ability of predicting precise bounding boxes learned in the source domain has been successfully transferred to the target domain. With our domain adaptation method to conduct better transferring, the improvement over the WSB is boosted to a higher level.
The above results shows each step in BR is necessary: the virtual scenes are helpful to boost the detection performance, and the domain adaptation method can further explore the potential of the virtual scenes. 
Interestingly, as the virtual scenes become more realistic (from point-version to mesh-version), the performance of BR improves a lot while WSBP has little change, which indicates that layout may not be that important in pretraining as in domain adaptation. 

\begin{table}
  \centering
  \setlength\tabcolsep{8pt}
  \caption{The detection results (mAP@0.25) of BR under different error rate for center labeling on ScanNet. We adopt GroupFree3D as the detector and utilize mesh-version virtual scenes for BR.}
  \vspace{-1mm}
  \begin{tabular}{c|ccccc}
      \hline
      \multirow{2}{*}{Method} & \multicolumn{5}{c}{Error Rate} \\
      \cline{2-6}
       & 10\% & 20\% & 30\% & 40\% & 50\% \\
      \hline
       WSB &29.7 &26.8 &25.0 &22.3 &19.7 \\
       BR$_M$(Ours) &\textbf{47.1} &\textbf{46.0} &\textbf{43.9} &\textbf{43.1} &\textbf{41.2}  \\
      \hline
  \end{tabular}
  \vspace{-2mm}
  \label{robust}
\end{table}

In terms of class-specific results, on some categories the mAP@0.25 of the BR$_M$ (for GroupFree3D) is even the highest among all the methods including the FSB. However, all methods fail to precisely detect cup and bottle, which shows current 3D detectors still face huge challenges in small object detection. More detection results (mAP@0.5) can be found in supplementary.

\textbf{Robustness for Labeling Error:}
In our labeling strategy, the center error is within 10\%, which we define as the error rate, of the object's size. To show the robustness of our approach, we gradually increase this rate from 10\% to 50\% by randomly jittering the centers according to the box sizes, and report the detection results of WSB and BR$_M$ (for GroupFree3D) in terms of mAP@0.25. As shown in Table \ref{robust}, with the increasing of error rate, the performance of BR degrades more slowly than WSB. Even if the error rate is 50\%, which allows us to label the centers in a more time-saving strategy, BR can still achieve satisfactory results (higher than 0.41 in terms of mAP@0.25).

\begin{figure}
  \centering
  \includegraphics[width=1.0\linewidth]{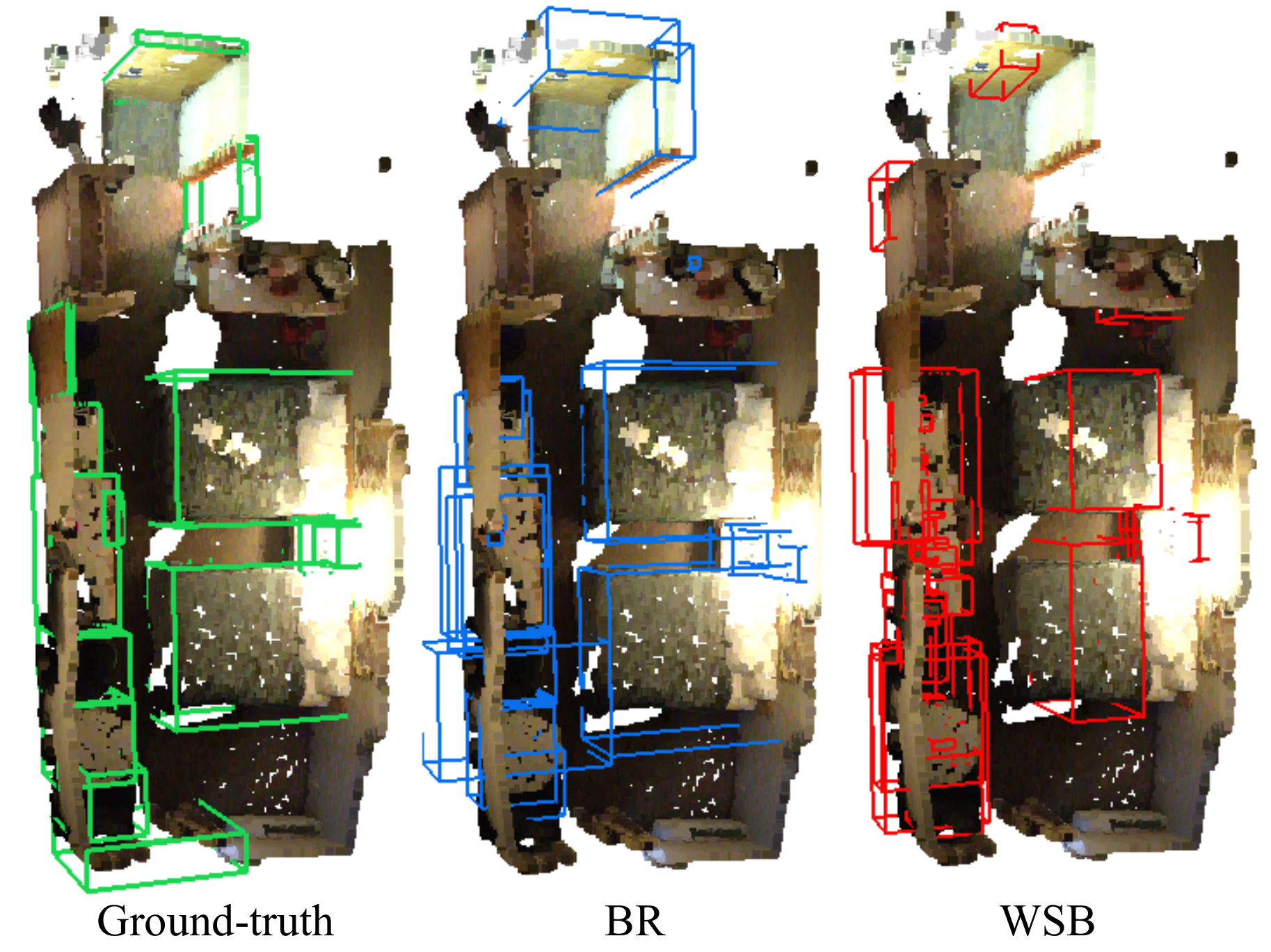}
  \vspace{-18pt}
  \caption{Visual Results on ScanNet. We compare BR and WSB with the ground-truth bounding boxes.}
  \vspace{-8pt}
  \label{fig:vis3}
\end{figure}

\textbf{Visualization Results:}
We visualize the detection results of WSB and BR$_M$ (for GroupFree3D) on ScanNet. As shown in Figure \ref{fig:vis3}, BR can produce more accurate detection results with less false positives. The visual results further confirm the effectiveness of the proposed method.

\subsection{Ablation Study}
We further design ablation experiments to study the influences of each scene generation step and each domain adaptation loss to the performance of our BR approach. In this section, we adopt VoteNet as the detector and use point-version virtual scenes for universality.




In Table \ref{tab4}, we illustrate that in our virtual scene generation pipeline, the physical constraints and density control are effective. As the virtual scenes become more realistic, the performance of our BR approach is getting better.

As shown in Table \ref{tab5}, we show the effect of each domain adaptation module and the center refine module. It can be seen that with global alignment or object proposal alignment, the detection performance can be boosted by 3.5\% and 2.2\% respectively. By combining the two kinds of feature alignments, we achieve higher detection accuracy. Having applied the center refinement method, the performance is further boosted by 1.0\%.

\subsection{Limitation}
Due to the limited number of categories in ModelNet40, we selectively evaluate the performance of BR on 22 classes. However, as online repositories of user-generated 3D shapes, such as the 3D Warehouse repository~\cite{Warehouse}, contain 3D shapes in almost any category, BR can be easily extended to 3D object detection on more classes once these online synthetic shapes are organized into a standard dataset. Therefore, ideally we can leverage a larger synthetic 3D shape dataset, which covers all objects that may appear in indoor scenes. This dataset can promote more researches on 3D scene understanding with synthetic shapes, which we leave for future work.

\begin{table}[]
  \centering
  \setlength\tabcolsep{7pt}
  \caption{The detection results (mAP@0.25) of BR with virtual scenes at different generation stages on ScanNet. Here the detector is VoteNet and the virtual scenes are point-version.}
  \vspace{-1mm}
  \begin{tabular}{p{1.5cm}<{\centering}p{1.5cm}<{\centering}p{1.5cm}<{\centering}|p{1.6cm}<{\centering}}
      \hline
      Gravity & Collision & Density & \multirow{2}{*}{mAP@0.25} \\
      Constrain & Constrain & Control \\
      \hline
       & & & 26.3 \\
       \checkmark & & & 27.2 \\
       \checkmark & \checkmark & & 28.5 \\
       \checkmark & \checkmark &\checkmark & \textbf{31.2} \\
      \hline
  \end{tabular}
  \label{tab4}
  \vspace{-1mm}
\end{table}

\begin{table}[]
        \centering
        \setlength\tabcolsep{7pt}
        \caption{The detection results (mAP@0.25) of BR with different domain adaptation modules on ScanNet. Here the detector is VoteNet and the virtual scenes are point-version.}
        \vspace{-1mm}
        \begin{tabular}{p{1.5cm}<{\centering}p{1.5cm}<{\centering}p{1.5cm}<{\centering}|p{1.6cm}<{\centering}}
            \hline
            Global & Proposal & Center  & \multirow{2}{*}{mAP@0.25} \\
            Alignment & Alignment & Refinement  \\
            \hline
             & & & 24.2 \\
             \checkmark & & & 28.7 \\
             & \checkmark & & 27.4 \\
             \checkmark & \checkmark & & 30.2 \\
             \checkmark & \checkmark & \checkmark & \textbf{31.2} \\
            \hline
        \end{tabular}
        \label{tab5}
        \vspace{-2mm}
\end{table}

\section{Conclusion}
In this paper, we have proposed a new label enhancement approach, namely Back to Reality (BR), for 3D object detection trained using only object centers and class tags as supervision.
To fully exploit the information contained in the position-level annotations, we consider them as the coarse layout of scenes, which is utilized to assemble 3D shapes into fully-annotated virtual scenes. We apply physical constraints on the generated virtual scenes to make sure the relationship between objects is reasonable. In order to make use of the virtual scenes to remedy the information loss from box annotations to centers, we present a virtual-to-real domain adaptation method, which transfers the useful knowledge learned from the virtual scenes to real-scene 3D object detection. Experimental results on ScanNet dataset show the effectiveness of our BR approach.


\section*{Acknowledgements}
This work was supported in part by the National Natural Science Foundation of China under Grant 62125603, and Grant U1813218, and in part by a grant from the Beijing Academy of Artificial Intelligence (BAAI).

{\small
\bibliographystyle{ieee_fullname}
\bibliography{egbib}
}
\appendix
\section*{Supplementary Material}
\section{overview}
\noindent This supplementary material\footnote{We include our code in the folder "BackToReality". Please refer to the README file for more details.} is organized as follows: 
\begin{itemize}
\itemsep0em 
    \item Section {\color{red}1} details the Approach section in the main paper. 
    \item Section {\color{red}2} shows the implementation detail of WS3D.
    \item Section {\color{red}3} details our augmentation strategy for small objects during training.
    \item Section {\color{red}4} shows more experimental results.
\end{itemize}

\section{Approach Details}
\label{sec:detail}

In this section, we show the details in our approach, which is divided into shape-guided \textbf{label enhancement} and virtual2real \textbf{domain adaptation}. 

\subsection{Label Enhancement}
We show the exact definitions of some concepts appeared in Section 3.2 of the main paper as below.

\textbf{Shape Properties:}
The $MER$ is computed in XY plane, which is the minimum rectangle enclosing all the points of the object template.
The $SSH$ is the height of the largest surface on which other objects can stand.
The $CSS$ is a boolean value, indicating whether the supporting surface is similar with the $MER$ (i.e. we can use the $MER$ to approximate the supporting surface if $CSS$ is true).

In order to calculate $MER$, we use the OpenCV~\cite{opencv.org} toolbox to calculate the $MER$ of 2D point set. As OpenCV cannot be directly utilized to process point clouds, we first project the object templates to XY plane to acquire 2D point sets. Then we calculate the $MER$ of a point set $S=\{(x_1, y_1), (x_2, y_2), ..., (x_n, y_n)\}$ as below:
\begin{align}
    (x, y, l, w, \theta)&={\rm minAreaRect}(1000*S) \\
    MER&=(x, y, \frac{l}{1000}, \frac{w}{1000},\theta)
\end{align}
where minAreaRect is a function in OpenCV, which takes integer 2D point set as input and returns a rectangle, and rectangle is represented by a quintuple $(x, y, length, width, \theta)$, which indicates the center coordinate, length, width and rotation angle of a rectangle. $1000*S$ means that we multiply all the coordinates in $S$ by 1000 and then convert the coordinates from float to integer, which can reduce the rounding error.

To compute $SSH$, we first utilize Open3D~\cite{open3d.org} to get the normals of each point from point cloud. Then if the normal of a point is almost vertical (i.e.\ the normal's length along Z-axis is greater than 0.88), we record the coordinate of this point. After traversing all the points, we have recorded a list of coordinates. We sort the list according to the Z coordinate in ascending order, and the list of sorted Z coordinate is named as $l_z$. Then get a slice of $l_z$ from index $\lfloor\frac{4}{5}len_z\rfloor$ to $\lfloor\frac{9}{10}len_z\rfloor$, where $len_z$ denotes the length of $l_z$. $SSH$ can be calculated by averaging this slice. Note that this algorithm suppose the supporter has a large supporting surface on its top, and it can tolerate $10\%$ points higher than this surface.

To calculate $CSS$, we collect points which satisfy $SSH-\frac{1}{10}h<z<SSH+\frac{1}{10}h$ from the given object template, where $h$ is the height of this object template. Then we project these points to XY plane and name them supporter points $P_S$. If $P_S$ can almost fill the $MER$, the $CSS$ is set to be $True$. To analyze the compactness, we use K-means algorithm to divide $P_S$ into 2 clusters: $P_{S1}$ and $P_{S2}$. Then we calculate the area of convex hull of $P_{S1}$ and $P_{S2}$. The area is computed by using OpenCV:
\begin{equation}\label{area}
    A=\frac{{\rm contourArea}({\rm convexHull}(1000*P))}{1000000}
\end{equation}
where contourArea and convexHull are functions in OpenCV, $P$ is a 2D point set and $A$ is the area of $P$. The areas for $P_{S1}$ and $P_{S2}$ are $A_1$ and $A_2$ respectively. So we can compute $CSS$ as below:
\begin{equation}
CSS=
\begin{cases}
True,& A_1+A_2>0.9*l*w\\
False,& otherwise
\end{cases}
\end{equation}
where $l$ and $w$ are the length and width of the $MER$ of this object template.

\textbf{Segment Properties:}
Next we provide the definitions of horizontal segment, the area of segment and the height of segment.

For a segment, we define $z$ as the Z coordinate of all the points on it. Then if $|maximum(z)-median(z)|<0.2$ or $|minimum(z)-median(z)|<0.2$, we consider this segment is horizontal. To calculate the area of segment, we directly utilize (\ref{area}) and take all points on the segment as input (ignore the Z coordinates of points). To compute the height of a segment, we follow the same procedure as computing $SSH$: we first calculate the normals and pick out points with normals that are almost vertical, and then we pick out the Z coordinates of these points and acquire a list $l_z$. The segment's height is defined as the mean of $l_z$.

\subsection{Domain Adaptation}
We first provide detailed definition of $L_3$. Then we show the architectures of our center refinement module and the two discriminators.

For weakly-supervised training, as only objects' centers and semantic classes are available, we set $L_3$ as a simpler version of $L_2$:
\begin{equation}
	L_3=L_f+L_i,\ L_f=L_s+L_o+L_c
\end{equation}
$L_f$ is used to supervise the final prediction, where $L_s$ and $L_o$ are the cross entropy losses for semantic labels and objectness scores, and $L_c$ is defined as:
\begin{equation}
	L_c=\sum_i{max(||C_{gi}-C_i||_2-\lambda S_{gi}, 0)}
\end{equation}
which denotes the hinge loss for centers. $C_i$ is the $i$-th predicted center, $C_{gi}$ is the nearest ground-truth center to $C_i$, and $S_{gi}$ indicates the average size for the semantic class of this object. We set $\lambda=0.05$ to approximate the labeling error of centers. For $L_i$, we only make use of the center coordinates to weakly supervise the intermediate process of training. For example, in VoteNet~\cite{Qi_2019_ICCV}, the detection module predicts votes from the semantic features and aggregate them to generate object proposals, in which voting coordinates are the intermediate variables need to be supervised. Here we utilize the Chamfer Distance between the voting coordinates and the ground-truth center coordinates to supervise the voting. In GroupFree3D~\cite{liu2021group}, the detection module utilize KPS to sample the semantic features and generate initial object proposals, where the sampled points require supervision. Originally the KPS operation requires us to sample the nearest k points to the object center from the point cloud belong to this object. However, we weaken this requirement and sample the nearest k points without any constraints.

For the center refinement module, we adopt the Set Abstraction (SA) layer~\cite{qi2017pointnet++} to extract feature from the local KNN graph. Then a MLP is utilized to predict center offset from the feature. The SA layer first concatenates the relative coordinates between the center and its neighbors to the features of the neighbors, which is followed by a shared MLP ($MLP(256,128)$\footnote{Numbers in bracket are output layer sizes. Batchnorm is used for all layers with ReLU except for the final prediction layer in $MLP(64,3)$.}) and a channel-wise max-pooling layer. The pooled feature contains the local information of the center, which is then concatenated with the one-hot vector of the center's semantic class (we name the feature after concatenation as center feature). We utilize another MLP ($MLP(64,3)$) to predict the center offset from the center feature.
\begin{figure}
	\centering
	\includegraphics[width=1.0\linewidth]{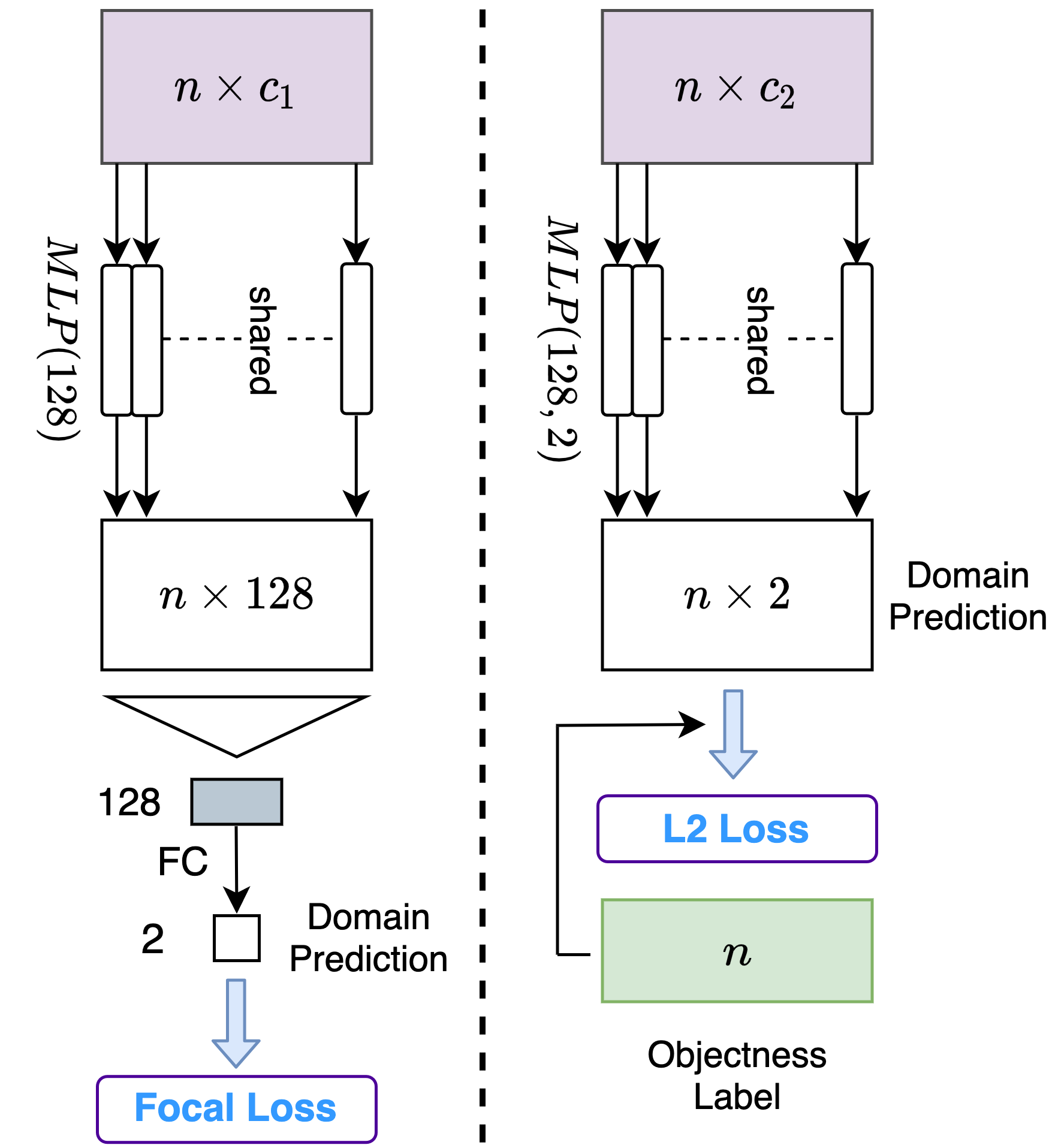}
	\vspace{-18pt}
	\caption{Architecture of the global and proposal discriminators. (Global on the left, proposal on the right.)}
	\vspace{-12pt}
	\label{fig:arch}
\end{figure}
For the global and proposal discriminators, we show their architectures in Figure \ref{fig:arch}.

\begin{table*}[]
	\centering
	\setlength{\abovedisplayskip}{0pt}
	\setlength{\belowdisplayskip}{0pt}
	\footnotesize
	\setlength\tabcolsep{1.6pt}
	\caption{The class-specific detection results (mAP@0.5) of different weakly-supervised methods on ScanNetV2 validation set. (FSB is the fully-supervised baseline. $^\dag$ indicates the method requires a small proportion of bounding boxes to refine the prediction. Other methods only use position-level annotations as supervision.)}
	\vspace{-1mm}
	\begin{tabular}{c|l|cccccccccccccccccccccc|c}
		\hline
		 & Setting &batht. &bed &bench & bsf. & bot. &chair & cup &curt. &desk &door & dres. & keyb. &lamp &lapt. & monit. & n.s. &plant & sofa & stool & table & toil. & ward. &mAP@0.5 \\
		\hline \hline
		\rowcolor{mygray}
		\cellcolor{white} \multirow{6}{*}{\rotatebox[origin=c]{90}{VoteNet}} & FSB~\cite{Qi_2019_ICCV} & 69.8 & 76.9 & 6.7 & 26.0 & 0.0 & 67.6 & 0.0 & 10.2 & 30.0 & 13.3 & 21.1 & 0.0 & 15.5 & 0.0 & 19.6 & 47.9 & 3.1 & 70.4 & 10.1 & 38.9 & 85.0 & 2.7 & 28.0 \\
		 & WSB & 0.0 & 11.5 & 0.0 & 0.0 & 0.0 & 1.7 & 0.0 & 0.0 & 0.0 & 0.0 & 0.0 & 0.0 & 0.6 & 0.0 & 0.2 & 0.7 & 0.0 & 0.2 & 0.0 & 0.1 & 2.5 & 0.0 & 0.8\\ 
		 & WS3D $^\dag$~\cite{meng2020weakly} & 0.0 & 22.7 & 0.0 & 0.0 & 0.0 & 12.2 & 0.1 & 0.0 & 0.3 & 0.0 & 0.0 & 0.0 & 1.1 & 0.0 & 1.3 & 11.3 & 0.0 & 0.1 & 0.2 & 1.4 & 16.4 & 0.0 & 3.1\\
		 & WSBP$_P$ & 0.0 & 3.7 & 0.0 & 0.1 & 0.0 & 28.4 & 0.0 & 0.0 & 1.1 & 0.5 & 0.0 & 0.0 & 1.2 & 0.0 & 0.0 & 26.1 & 0.0 & 0.9 & 4.4 & 0.8 & 7.6 & 0.6 & 3.4\\ %
		 & WSBP$_M$ & 12.3 & 1.3 & 0.0 & 0.3 & 0.0 & 16.5 & 0.0 & 0.0 & 4.1 & 0.1 & 0.0 & 0.4 & 5.9 & 0.0 & 0.1 & 26.9 & 0.4 & 3.3 & 5.4 & 0.8 & 4.8 & 0.1 & 3.8\\
		 & BR$_P$(Ours) & 36.8 &15.2 &1.2 &6.9 &0.0 &42.7 &0.0 &0.0 &4.4 &1.3 &2.1 &0.0 &9.0 &0.0 &2.7 &31.4 &1.3 &14.4 &4.1 &8.3 &51.6 &0.0 &10.6 \\
		 & BR$_M$(Ours) & 9.6 & 59.2 & 0.2 & 12.8 & 0.0 & 37.9 & 0.0 & 0.0 & 22.1 & 1.0 & 6.2 & 0.0 & 10.6 & 0.0 & 2.1 & 44.6 & 2.7 & 33.0 & 2.0 & 25.3 & 57.0 & 0.1 & 14.8\\
		\hline
		\rowcolor{mygray}
		\cellcolor{white} \multirow{6}{*}{\rotatebox[origin=c]{90}{GroupFree3D}} & FSB~\cite{liu2021group} & 75.7 & 75.6 & 4.5 & 28.4 & 0.0 & 75.3 & 0.0 & 20.3 & 47.4 & 24.7 & 29.5 & 0.3 & 20.4 & 0.0 & 37.5 & 61.4 & 3.7 & 74.6 & 37.1 & 51.1 & 96.2 & 11.7 & 35.2\\
		 & WSB & 1.9 & 24.7 & 0.0 & 0.1 & 0.0 & 31.2 & 0.0 & 0.0 & 0.1 & 0.1 & 0.0 & 0.0 & 6.5 & 0.0 & 2.1 & 1.5 & 0.1 & 2.6 & 2.0 & 0.5 & 54.3 & 0.0 & 5.8\\ 
		 & WS3D$^\dag$~\cite{meng2020weakly} & 3.8 & 25.7 & 0.0 & 0.1 & 0.0 & 36.4 & 0.0 & 0.0 & 2.1 & 0.0 & 0.3 & 0.3 & 10.2 & 0.0 & 7.5 & 16.4 & 0.2 & 2.7 & 4.5 & 0.4 & 68.3 & 0.0 & 8.1\\
		 & WSBP$_P$ & 1.9 & 5.2 & 0.0 & 1.3 & 0.0 & 31.8 & 0.0 & 11.3 & 1.1 & 0.1 & 0.0 & 0.0 & 18.7 & 4.4 & 1.0 & 48.1 & 1.3 & 1.3 & 1.3 & 0.6 & 62.0 & 1.8 & 8.3\\ %
		 & WSBP$_M$ & 4.9 & 16.7 & 0.0 & 0.5 & 0.0 & 34.1 & 0.0 & 0.1 & 5.6 & 0.2 & 0.5 & 0.1 & 9.0 & 4.6 & 8.9 & 48.5 & 0.9 & 9.9 & 12.3 & 3.4 & 51.9 & 0.0 & 9.6\\
		 & BR$_P$(Ours) & 83.6 & 79.1 & 0.0 & 10.8 & 0.0 & 53.5 & 0.0 & 0.0 & 0.0 & 1.6 & 3.7 & 0.0 & 19.6 & 50.0 & 6.5 & 60.0 & 16.7 & 21.1 & 5.7 & 14.6 & 90.1 & 0.0 & 23.5\\
		 & BR$_M$(Ours) & 83.3 & 65.0 & 0.0 & 4.1 & 0.0 & 56.2 & 0.0 & 0.5 & 11.8 & 2.1 & 16.7 & 1.2 & 23.8 & 12.5 & 16.0 & 80.0 & 17.5 & 42.2 & 28.6 & 28.0 & 99.2 & 0.0 & 26.8\\
		\hline
	   \end{tabular}
	\label{tab1}
	\vspace{-2mm}
\end{table*}

\section{Implementation Detail of WS3D}
\label{sec:ws3d}
In this section, we show how we implement WS3D~\cite{meng2020weakly} to adapt to indoor 3D object detection task.

\subsection{Introduction of WS3D}
Here is a simple summary of WS3D: The authors annotate the object centers in the bird's eye view (BEV) maps, which takes 2.5s per object. Then they utilize a two-stage approach to detect a specific category of objects (the author focus on \textbf{Car} in their paper), which can be divided into proposal and refinement stages.
At the proposal stage, WS3D creates cylindrical proposals from the labeled centers, whose radius and height are fixed since the sizes of cars are close. Therefore the probability of a car being wrapped in a cylindrical proposal is high. Then a network (Net1) is trained to generate proposals from a point cloud scene.
At the refinement stage, another network (Net2) is trained to take in the cylindrical proposal and output the bounding box of the car contained in the proposal, where around 3\% well-labeled instances are used for supervision.

\subsection{Proposal Stage}

Since the indoor scenes in ScanNetV2 are more complicated, the size and height of each object is different, even for objects in the same class. Therefore we annotate the object centers in 3D space rather than in the BEV map, which is the same labeling strategy with us and takes 5s per object, to provide stronger supervision for WS3D. 
Instead of using a simple fixed-size cylinder as the proposal, we utilize a cuboid instead, whose size (length, width and height) is 1.5 times the average size of the object's category. In this way we are able to generate a more reasonable proposal.

During this stage, we can adopt different detectors as Net1. Net1 is trained with position-level annotations and used to predict the centers and semantic labels of objects (we adopt VoteNet and GroupFree3D as Net1 in our experiments). Then we generate cuboid proposals from the predicted centers and classes.

\subsection{Refinement Stage}

We find 3\% well-labeled bounding boxes are not enough to train the Net2, as there 22 categories in our benchmark and the size of each object is very different, so we use around 15\% bounding boxes instead. The proposals generated from the previous stage are post-processed by a 3D NMS module with an IoU threshold of 0.25, and then refined into precise bounding boxes by Net2.

We adopt a PointNet++-like module as Net2, whose input is the point cloud inside the cuboid proposal and output is the refined center coordinate, box size and box orientation.

\section{Augmentation Strategy}
\label{sec:aug}
As the number of scenes which contain small objects\footnote{Small objects are \{bottle, cup, keyboard\}.} and the probability of small objects being sampled are relatively smaller than others, it is difficult for the detector to learn how to locate small objects in complex scenes. Therefore we utilize an augmentation strategy similar to \cite{kisantal2019augmentation} to handle the problem.

During trianing, we oversample the virtual scenes which contain small objects twice in each epoch. We further copy-paste small objects to the oversampled virtual scenes: for each small object, we copy it with a probability of 0.75 and paste it randomly in the scene (the pasted center must be in the axis-aligned bounding box of the whole scene). Then we apply gravity and collision contraints and control the densities of these added small objects as mentioned in the virtual scene generation method.

Apart from small objects, we also consider the scarce objects\footnote{Scarce objects are \{bathtub, bench, dresser, laptop, wardrobe\}.}, as the number of them is relatively small and thus the detector is not sufficiently trained on these categories. We add the scarce objects to the oversampled virtual scenes to expand the number of them. We first decide how many objects of each scarce category we should add according to Table 2 in the main paper, where we set 40, 70, 15, 55 and 50 for bathtub, bench, dresser, laptop and wardrobe respectively. Then we choose scenes which are suitable for adding these objects by calculating the value of correlation between scenes and scarce categories as below:
\begin{equation}
    Corr(s, c)=\sum_{i=1}^{22}{l_{s_i}(v_{c_i}-r)}
\end{equation}
where $s$ indicates a scene and $c$ denotes a scarce category. $l_s$ is a 22-dimensional boolean vector where $l_{s_i}$ indicates whether there is an object of the $i$-th category in $s$. $v_c$ is a 22-dimensional vector which indicates the correlation between $c$ and other categories:
\begin{equation}
    v_{c_i}=
    \begin{cases}
        \frac{Num(i, Index(c))}{Num(Index(c))},& i \neq Index(c) \\
        0,& i = Index(c)
    \end{cases}
\end{equation}
where $Num(...)$ is a function, whose input is a set of indexs of category and output is the number of scenes which contain objects in all the input categories. The larger $v_{c_i}$, the stronger the correlation between $c$ and the $i$-th category. As we hope the highly correlated scenes for $c$ do not contain too many categories with low $v_{c_i}$, we introduce a penalty term $r$ to reduce the value of $Corr(s, c)$ when there are a large number of categories weakly correlated to $c$ in $s$. We set $r=0.25$ in our experiments.

\section{More Detection Results}
\label{sec:exp}
We show 3D object detection results (mAP@0.5) of different weakly-supervised methods on ScanNetV2~\cite{dai2017scannet} validation set in Table \ref{tab1}.

Consistent with the results on mAP@0.25, our BR approach achieves the best performance among all the weakly-supervised approaches. Under a more strict metric, the performances of most weakly-supervised approaches fail to surpass 10\% in terms of mAP@0.5, that shows it is really hard to precisely detect the objects in a complicated indoor scene for a detector trained with only position-level annotations. However, the performance of BR$_M$ (for GroupFree3D) still achieves 26.8\% in terms of mAP@0.5, which is comparable to the performance of fully-supervised VoteNet.

We also find our BR approach works better on GroupFree3D than on VoteNet (the gap between FSB and BR is smaller). This may be due to the features extracted by stronger detector has better generalization ability and thus our virtual2real domain adaptation method can transfer more useful knowledge contained in the virtual scenes to real-scene training.

\end{document}